\documentclass{article}

\usepackage[final, nonatbib]{neurips_2024}

\usepackage[utf8]{inputenc} 
\usepackage[T1]{fontenc}    
\usepackage{booktabs}       
\usepackage{amsfonts}       
\usepackage{nicefrac}       
\usepackage{microtype}      

\usepackage[table,xcdraw]{xcolor} 

\usepackage{url}

\usepackage{breakurl}
\usepackage[breaklinks]{hyperref}

\usepackage{amsmath}
\usepackage{graphicx}
\usepackage{threeparttable}
\usepackage{wrapfig}
\usepackage[normalem]{ulem} 
\useunder{\uline}{\ul}{}
\usepackage{placeins}

\title{Tokenization Standards for Linguistic Integrity: Turkish as a Benchmark}

\author{
M. Ali Bayram$^1$, Ali Arda Fincan$^2$, Ahmet Semih Gümüş$^2$, Sercan Karakaş$^3$, \\
Banu Diri$^1$, Savaş Yıldırım$^4$\\
$^1$Yıldız Technical University, $^2$Yeditepe University, $^3$University of Chicago, \\
$^4$Istanbul Bilgi University \\
\texttt{malibayram20@gmail.com}
}

\begin{document}
\maketitle

\begin{abstract}
Tokenization constitutes an essential preprocessing step within Natural Language Processing (NLP), exerting a direct impact on large language models’ (LLMs) capacity to capture syntactic, morphosyntactic, and semantic details. This paper introduces a novel framework for the systematic evaluation of tokenization strategies, with a particular focus on mitigating the challenges associated with morphologically-rich and low-resource languages. Using a Turkish dataset of 6,200 multiple-choice questions derived from the Massive Multitask Language Understanding (MMLU) benchmark, the framework evaluates tokenizers across five key metrics: vocabulary size, token count, processing time, \textit{language-specific token percentages} (\%TR), and \textit{token purity}. These metrics, proposed in this study, offer a structured approach to assessing how effectively tokenizers preserve linguistic structures. While \%TR measures the proportion of valid words generated in the target language, \%Pure evaluates the alignment of tokens with meaningful linguistic units, such as roots and valid morphemes, ensuring minimal semantic fragmentation. The findings reveal that \textit{language-specific token percentages}, introduced as a critical evaluation metric, exhibit a stronger correlation with downstream performance (e.g., MMLU scores) compared to token purity, emphasizing their importance in enhancing model accuracy and robustness. Furthermore, the analysis demonstrates that larger model parameters do not necessarily yield superior tokenization quality or improved results, highlighting the need for tailored tokenization strategies that prioritize linguistic alignment over mere computational scaling. By addressing both computational efficiency and linguistic fidelity, this framework establishes a new standard for developing robust tokenization methods optimized for morphologically complex and low-resource languages. Future work will focus on advancing morphological analysis techniques, exploring domain-specific customizations, and conducting cross-linguistic evaluations to further refine tokenization practices across diverse linguistic contexts.

\end{abstract}
\section{Introduction}

Tokenization is a crucial preprocessing step in NLP, transforming raw text into smaller units such as words, subwords, or characters that serve as inputs for language models. This process directly impacts the efficiency and effectiveness of NLP tasks. While tokenization is a universal requirement across languages, its complexity increases for morphologically rich and agglutinative languages like Turkish, where words often consist of a root morpheme and multiple morphemes, each carrying distinct grammatical or semantic meaning \cite{schmidt_tokenization_2024}.

Recent research has explored various tokenization methods to address these challenges. Domingo et al.~\cite{domingo2019how} and Fujii et al.~\cite{fujii2023how} have highlighted the significant influence of tokenization on downstream tasks such as neural machine translation and scriptio continua languages, respectively. Theoretical studies by Guo~\cite{guo1997critical} and Kudo and Richardson~\cite{kudo_sentencepiece_2018} examined the foundational principles of effective tokenization, providing the groundwork for widely used methods like Byte-Pair Encoding (BPE) and SentencePiece. Additionally, Zouhar et al.~\cite{zouhar2023formal} emphasized the importance of balancing vocabulary size and segmentation granularity, while Berglund and van der Merwe~\cite{berglund_formalizing_2023} provided a formal framework for understanding BPE tokenization semantics in modern NLP systems. These contributions collectively advance the understanding of tokenization's impact across diverse NLP tasks.

Tokenization challenges are amplified for non-English languages due to increased token fragmentation. Tokenizers often produce smaller chunks for non-English texts, significantly increasing token counts and bloating sequence lengths. This inefficiency wastes the context window of language models, degrading performance for non-English queries. These issues are closely tied to tokenizer design and the training data used for their development \cite{samiullah_technical_nodate}.

Recent advancements in subword tokenization techniques, such as Byte Pair Encoding (BPE) \cite{gage_new_1994} and SentencePiece \cite{kudo_sentencepiece_2018}, have demonstrated improvements in representing complex linguistic structures across languages. These methods segment words into subword units, enabling models to handle rare and unseen words more effectively. For instance, the \texttt{Aranizer-BPE-86k} tokenizer, developed for Arabic, effectively captures the morphological nuances of the language, offering insights for handling similar challenges in Turkish \cite{koubaa_githubcomriotu-labaranizer_2024}. These methods are also relevant for morphologically simpler languages, such as English, where they enhance representation efficiency, particularly in data-scarce scenarios.

\textbf{Motivational Example:} Consider the Turkish word \textit{"evlerimizden"} (\textit{"from our houses"}). A naive tokenizer might split it into segments like [\texttt{"ev"}, \texttt{"ler"}, \texttt{"imiz"}, \texttt{"den"}], which captures the morphemes roughly, while a linguistically uninformed tokenizer might produce unaligned fragments such as [\texttt{"e"}, \texttt{"vl"}, \texttt{"er"}, \texttt{"imizd"}, \texttt{"en"}], disrupting semantic coherence. While \texttt{"evler"} is a valid Turkish word and contributes to language-specific token percentages, it contains multiple meaningful morphemes and thus reduces token purity. Tokens like \texttt{"imiz"} or \texttt{"den"} are neither valid Turkish words nor linguistically pure tokens, highlighting the challenges of balancing these metrics.

Even in languages with simpler morphologies, such as English, tokenization can introduce challenges. For example, a naive tokenizer might incorrectly split the compound word \textit{"unbelievable"} into fragments like [\texttt{"un"}, \texttt{"belie"}, \texttt{"vable"}], losing semantic coherence and structure. A linguistically informed tokenizer would segment it meaningfully as [\texttt{"un"}, \texttt{"believe"}, \texttt{"able"}], preserving linguistic integrity and improving downstream performance.

Two critical metrics proposed in this study for evaluating tokenization quality are \textit{token purity} and \textit{language-specific token percentages}, offering complementary insights into the effectiveness of tokenization strategies.

\textbf{Token Purity} measures how well the generated tokens align with meaningful linguistic units, such as roots, valid morphemes, or coherent semantic segments. High token purity minimizes unnecessary fragmentation of words and ensures that the tokens retain their linguistic integrity. For instance, segmenting the Turkish word \textit{"evlerimizden"} into tokens like [\texttt{"ev"}, \texttt{"ler"}, \texttt{"imiz"}, \texttt{"den"}] achieves a certain level of purity, as it preserves the morphological components. However, tokens like [\texttt{"evler"}], while valid words in Turkish, are not considered pure because they contain multiple morphemes that could be further segmented into meaningful linguistic units. On the other hand, meaningless splits like [\texttt{"e"}, \texttt{"vl"}, \texttt{"eri"}, \texttt{"mizden"}] further reduce purity by fragmenting morphemes incoherently.

\textbf{Language-Specific Token Percentages} evaluate the proportion of generated tokens that are valid words in the target language. While token purity focuses on semantic and grammatical coherence within a word, language-specific token percentages assess whether tokens align with the vocabulary of the language, regardless of whether they represent finer linguistic units. For instance, a tokenizer that produces valid Turkish words like [\texttt{"ev"} and \texttt{"ler"}] contributes to a higher language-specific token percentage. Conversely, invalid tokens like [\texttt{"evlerd"}] or non-words like [\texttt{"imiz"}] negatively affect this metric because they are not recognized as valid words in Turkish.

These two metrics, while distinct, highlight different aspects of tokenization quality. High token purity ensures that meaningful parts of words are preserved, aiding linguistic understanding and model learning. High language-specific token percentages indicate that the tokens align with the vocabulary of the language, even if they do not capture finer-grained morphological units. However, language-specific tokens may still introduce noise by containing multiple meaningful morphemes, which could obscure semantic relationships and degrade model interpretability.

Despite advancements, achieving a balance between tokenization speed, vocabulary size, and linguistic fidelity remains challenging. Excessive fragmentation dilutes semantic meaning, while overly coarse tokenization overlooks critical linguistic details. This balance is essential for morphologically rich languages like Turkish and for optimizing performance in morphologically simpler languages \cite{neubeck_so_2024}.

This study evaluates tokenizers for Turkish using the MMLU benchmark, a widely recognized evaluation suite for language models. By analyzing tokenizers based on token purity, language-specific token percentages, vocabulary size, and processing speed, this work identifies effective approaches for Turkish NLP tasks. The findings contribute to optimized tokenization strategies that benefit diverse linguistic settings, advancing the accuracy and efficiency of NLP models.

\section{Related Work}

Tokenization plays a fundamental role in (NLP), directly influencing the performance, efficiency, and accuracy of (LLMs). Recent research has explored various tokenization strategies and their downstream impacts, aiming to balance linguistic fidelity, computational efficiency, and model scalability.

The \textit{Arabic Tokenizers Leaderboard} \cite{rashad_arabic_nodate} benchmarks tokenizers for Arabic, using datasets such as \texttt{rasaif-translations} and \texttt{Moroccan Arabic Wikipedia}, highlighting the unique challenges posed by Arabic's diverse dialects and orthographic complexity. Tools like \textit{AraNizer} \cite{koubaa_githubcomriotu-labaranizer_2024} leverage subword-based techniques, such as Byte Pair Encoding (BPE) and SentencePiece, to better capture the morphological nuances of Arabic and enhance downstream performance.

Similarly, the \textit{NbAiLab Tokenizer Benchmark} \cite{rosa_nbailabtokenizer-benchmark_2024} evaluates tokenization strategies for Scandinavian languages, emphasizing the critical need for language-specific adaptations in multilingual contexts. For German, Diewald et al. \cite{diewald_tokenizing_2022} assessed tokenizers like \texttt{KorAP-Tokenizer} and \texttt{SoMaJo}, achieving high accuracy in token boundary detection while ensuring computational efficiency for large-scale corpora.

Erkaya \cite{erkaya_analysis_2023} provides a comprehensive analysis of subword tokenization methods, particularly focusing on their application to morphologically rich languages like Turkish. Erkaya evaluates the impact of corpus size and vocabulary size on tokenization characteristics, highlighting that larger corpora improve morphology encoding. The study also introduces a morphologically optimized tokenizer that improves downstream performance on tasks such as named-entity recognition, parts-of-speech tagging, question answering, and sentiment analysis. This work emphasizes the significance of incorporating morphological supervision into tokenization for languages with agglutinative structures.

A significant contribution to multilingual tokenization comes from the EuroLLM team, which emphasizes the importance of designing tokenizers with large vocabularies to support diverse linguistic structures \cite{martins_eurollm_2024}. By employing a Byte Pair Encoding (BPE) tokenizer with byte fallback and a vocabulary of 128,000 pieces, EuroLLM achieves a balance between low fertility (tokens per word) and parameter efficiency. Their findings indicate that vocabulary size is a critical factor in determining a tokenizer’s ability to efficiently process multiple languages, including European and non-European ones. EuroLLM's comparison of fertility metrics across tokenizers, such as those from Mistral, LLaMA-3, and Gemma, further underscores the trade-offs between large vocabularies and computational cost.

Efficiency advancements have also been demonstrated in GitHub's faster BPE implementation \cite{neubeck_so_2024}, which significantly improves scalability for tasks requiring billions of tokens. This aligns with EuroLLM’s approach of optimizing tokenization to enhance downstream task performance while maintaining computational efficiency.

Rust et al. \cite{rust_how_2021} highlight the effectiveness of monolingual tokenizers tailored to specific languages, showing notable downstream improvements for morphologically rich languages. Similarly, Lin et al. \cite{lin_not_nodate} propose Selective Language Modeling (SLM), which assigns utility scores to tokens and selectively trains on high-utility tokens, reducing noise and enhancing training efficiency. This approach is particularly relevant for languages like Turkish, where preserving meaningful tokens is essential for capturing linguistic richness.

The studies discussed collectively emphasize the necessity of tokenization strategies that balance linguistic integrity, computational efficiency, and downstream performance. Building on these advancements, this study evaluates tokenizers for Turkish, employing metrics such as token purity, vocabulary size, and processing speed. By integrating insights from multilingual projects like EuroLLM and tailoring techniques for morphologically rich languages, this work advances the understanding and optimization of tokenization for diverse linguistic contexts.
\section{Methodology}

This study evaluates tokenization strategies for morphologically rich and agglutinative languages, with Turkish chosen as a representative case. While Turkish serves as the primary focus, the methodology is designed to be flexible and adaptable to other languages with similar tokenization challenges, such as Finnish, Hungarian, and Uyghur.

The evaluation employs the Turkish MMLU (TR-MMLU) dataset \cite{bayram_setting_2025}, a meticulously designed benchmark for evaluating the linguistic and conceptual capabilities of large language models (LLMs) in Turkish. The dataset comprises 6,200 multiple-choice questions across 62 sections, drawn from a pool of 280,000 questions that span 67 disciplines and over 800 topics within the Turkish education system. TR-MMLU covers diverse subject areas, including law, healthcare, history, and natural sciences, ensuring a comprehensive representation of Turkish linguistic structures. This benchmark provides a culturally and linguistically relevant framework that avoids translation-related errors and reflects the unique morphological and syntactic complexities of Turkish.

The questions in TR-MMLU were sourced from standardized exams such as the University Entrance Examination and the Open Education Faculty (AUZEF) exams, which are designed to evaluate not only factual recall but also conceptual understanding, logical reasoning, and contextual knowledge. For example, a typical question might ask: \textit{"Hangi organ karaciğerin görevini destekler?"} (\textit{"Which organ supports the function of the liver?"}) with options like (A) \textit{Kalp} (\textit{Heart}), (B) \textit{Akciğer} (\textit{Lung}), (C) \textit{Böbrek} (\textit{Kidney}), and (D) \textit{Dalak} (\textit{Spleen}). Such questions test the model's ability to process Turkish text holistically, including its semantic, syntactic, and contextual dimensions.

This dataset is natively crafted by experts in the Turkish education system, ensuring alignment with cultural and linguistic norms while eliminating errors typically introduced through translation-based benchmarks. Additionally, TR-MMLU excludes questions likely to overlap with pretraining datasets, providing an unbiased evaluation of model performance. The inclusion of questions that require nuanced understanding and reasoning makes TR-MMLU particularly effective for evaluating linguistic alignment and conceptual comprehension.

The same framework can be extended to other languages by adapting analogous linguistic resources and educational content. For example, languages like Finnish, Hungarian, or Uyghur, which share morphological richness and agglutinative structures with Turkish, can benefit from a similar approach. The flexibility of the methodology allows researchers to evaluate tokenization strategies and LLM performance in a manner that respects the linguistic intricacies of different languages.

Several metrics are used to evaluate both computational and linguistic aspects of tokenization.

\textbf{Vocabulary Size:}  
Vocabulary size represents the total number of unique tokens, such as words, subwords, or characters, that a tokenizer can produce. For instance, a tokenizer with a vocabulary size of 50,000 might include tokens like \texttt{"run"}, \texttt{"ning"}, \texttt{"play"}, and \texttt{"ful"} to represent words such as \texttt{"running"}, \texttt{"playful"}, or \texttt{"played"}. A tokenizer with a smaller vocabulary might split \texttt{"unbelievable"} into [\texttt{"un"}, \texttt{"belie"}, \texttt{"vable"}], while a larger vocabulary might tokenize it as [\texttt{"un"}, \texttt{"believable"}], preserving linguistic coherence. Larger vocabularies allow for capturing longer word sequences, but they also increase memory usage and computational complexity. Conversely, smaller vocabularies may lead to excessive fragmentation, reducing linguistic and semantic interpretability.

\textbf{Total Token Count:}  
Total token count measures the number of tokens generated when a tokenizer processes a dataset. Consider the Turkish sentence \texttt{"Çocuklar bahçede oynayacak ve bahçede gülecek"} (\texttt{"The children will play in the garden and will laugh in the garden"}). A space-based tokenizer would yield eight tokens: [\texttt{"Çocuklar"}, \texttt{"bahçede"}, \texttt{"oynayacak"}, \texttt{"ve"}, \texttt{"bahçede"}, \texttt{"gülecek"}]. However, a subword-based tokenizer might generate: [\texttt{"Çocuk"}, \texttt{"lar"}, \texttt{"bahçe"}, \texttt{"de"}, \texttt{"oyna"}, \texttt{"yacak"}, \texttt{"ve"}, \texttt{"bahçe"}, \texttt{"de"}, \texttt{"gül"}, \texttt{"ecek"}]. This illustrates the effects of repeated tokens (\texttt{"bahçe"} and \texttt{"de"}) on total token count. Lower total token counts can improve computational efficiency but may overlook detailed morphological structures.

\textbf{Processing Time:}  
Processing time, measured in seconds, indicates the computational efficiency of the tokenizer. For example, tokenizing a dataset with one million words might take 2.5 seconds. For a Turkish example, processing the sentence \texttt{"Çocuklar bahçede oynayacak ve bahçede gülecek"} with a simple tokenizer may take 0.1 seconds, while a more complex tokenizer accounting for subword or morphological structures might take 0.3 seconds. Faster tokenization is crucial for real-time applications, whereas computationally intensive tokenization methods may better preserve linguistic nuances.

\textbf{Language-Specific Token Percentages (\%TR):}  
This metric evaluates the proportion of unique tokens that are valid words in the target language. It reflects the tokenizer's ability to produce linguistically valid tokens, regardless of whether they can be further decomposed. For instance, the Turkish sentence \texttt{"Çocuklar bahçede oynayacak ve bahçede gülecek"} might be tokenized as [\texttt{"Çocuklar"}, \texttt{"bahçe"}, \texttt{"de"}, \texttt{"oynayacak"}, \texttt{"ve"}, \texttt{"gül"}, \texttt{"ecek"}]. Here, \texttt{"Çocuklar"}, \texttt{"bahçe"}, , \texttt{"oynayacak"}, \texttt{"ve"}, and \texttt{"gül"} are valid Turkish words, contributing to a higher \%TR, while \texttt{"de"} and \texttt{"ecek"} are not standalone Turkish words and do not contribute. If five out of seven unique tokens are valid Turkish words, \%TR is calculated as:

\begin{equation}
\%TR = \frac{\text{Valid Unique Tokens}}{\text{Unique Tokens}} \times 100
\label{eq:tr_percentage}
\end{equation}

Substituting the values yields:
\[
\%TR = \frac{5}{7} \times 100 = 71.4\%.
\]

This metric ensures that the tokenizer aligns with the vocabulary of the language but does not evaluate token granularity or linguistic purity.

\textbf{Pure Token Percentage (\%Pure):}  
This metric measures the proportion of unique tokens that are semantically pure, meaning they cannot be further decomposed into smaller meaningful linguistic units. Using the same sentence \texttt{"Çocuklar bahçede oynayacak ve bahçede gülecek"}, the tokenizer might generate unique tokens such as [\texttt{"Çocuklar"}, \texttt{"bahçe"}, \texttt{"de"}, \texttt{"oynayacak"}, \texttt{"ve"}, \texttt{"gül"}, \texttt{"ecek"}]. Among these, \texttt{"bahçe"}, \texttt{"gül"}, and \texttt{"ve"} are considered pure because they represent atomic linguistic units in Turkish. In contrast, \texttt{"Çocuklar"} and \texttt{"oynayacak"} are not pure, as they combine multiple morphemes. If three out of 7 unique tokens are pure, \%Pure is calculated as:

\begin{equation}
\%Pure = \frac{\text{Pure Unique Tokens}}{\text{Unique Tokens}} \times 100
\label{eq:pure_percentage}
\end{equation}

Substituting the values:
\[
\%Pure = \frac{3}{7} \times 100 \approx 42.9\%.
\]

While \%TR evaluates a tokenizer's ability to generate valid words, \%Pure emphasizes preserving linguistic granularity and atomicity. A tokenizer achieving high \%TR by generating valid tokens like \texttt{"Çocuklar"} might sacrifice \%Pure if these tokens combine multiple linguistic units.

By evaluating these metrics together, this study provides a comprehensive framework for assessing tokenization strategies. High \%TR values indicate alignment with the target language’s vocabulary, while high \%Pure values reflect the semantic and grammatical integrity of the tokens. Balancing these metrics is crucial for morphologically rich languages, where both linguistic alignment and token granularity significantly impact downstream NLP tasks.

These metrics, defined in Equations \ref{eq:tr_percentage} and \ref{eq:pure_percentage}, allow for a detailed quantitative evaluation of tokenization quality, highlighting trade-offs between computational efficiency, linguistic alignment, and semantic granularity.

Morphological analysis and token validation are performed using language-specific tools, including the ITU Turkish NLP Web Service \cite{eryigit_itu_2014} and the Kalbur library \cite{aksoy_ahmetaxkalbur_2024}. These tools identify valid roots, morphemes, and semantically coherent units, ensuring precise evaluations of tokenization quality. For other languages, similar linguistic analyzers and rule-based systems can be employed.

Computational metrics, such as processing time and token counts, are derived using Python scripts and the Hugging Face Tokenizers library \cite{neubeck_so_2024}. To ensure reproducibility, all datasets, scripts, and configurations are documented and made publicly accessible in a GitHub repository \cite{bayram_malibayramtokenizer_benchmark_2024}. This repository includes Python scripts, experimental details, and resources necessary for replicating the analyses. While Turkish is the benchmark language, the framework is adaptable to other languages and datasets, offering a versatile and scalable approach for evaluating tokenization strategies across diverse linguistic settings.
\section{Results and Analysis}

This study evaluated four state-of-the-art tokenizers using the TR-MMLU dataset, which consists of 1,605,376 characters and 198,193 words \cite{bayram_setting_2025}. The TR-MMLU benchmark provides a robust framework for assessing large language models across a broad spectrum of subjects, accurately capturing the unique morphological and syntactic structures of Turkish. The evaluation considered key metrics, including Turkish Token Percentage (TR \%), Pure Token Percentage (Pure \%), vocabulary size, total token count, unique token count, processing time, and MMLU scores. These metrics collectively capture both linguistic fidelity and computational efficiency.

\begin{table}[h]
\centering
\caption{Tokenizer Benchmark Results}
\label{tab:tokenizer-benchmark}
\resizebox{\textwidth}{!}{
\begin{tabular}{|l|c|c|c|c|}
\hline
\rowcolor[HTML]{DDDDDD} 
\textbf{Metric} & \textbf{gemma-2} & \textbf{llama-3.1} & \textbf{Qwen2.5} & \textbf{aya-expanse} \\ \hline
\rowcolor[HTML]{FFFFDD} 
Model Parameters (B) & 27.2 & 70.6 & 7.6 & 32.3 \\ \hline
\rowcolor[HTML]{FFFFDD} 
MMLU Score (\%) & 72.10 & 70.42 & 61.68 & 70.66 \\ \hline
Vocabulary Size & 256,000 & 128,256 & 151,665 & 255,029 \\ \hline
Token Count & 497,015 & 488,535 & 561,866 & 434,526 \\ \hline
Processing Time (s) & 2.95 & 3.12 & 3.31 & 2.77 \\ \hline
Unique Token Count & 6,383 & 6,823 & 5,752 & 8,562 \\ \hline
TR \% & 48.63 & 45.80 & 40.33 & 50.67 \\ \hline
Pure \% & 37.05 & 30.91 & 30.15 & 32.96 \\ \hline
\end{tabular}
}
\end{table}

Table~\ref{tab:tokenizer-benchmark} summarizes the performance metrics for the evaluated tokenizers. Among the models, \texttt{gemma-2} achieved the highest MMLU score (72.10\%) and the highest Pure \% (37.05\%), demonstrating its superior ability to generate linguistically coherent tokens while maintaining strong downstream performance. Its TR \% (48.63\%) further reflects its alignment with Turkish vocabulary and morphological structures.

\texttt{aya-expanse} recorded the highest TR \% (50.67\%) and performed competitively on the MMLU benchmark with a score of 70.66\%. This result highlights the significance of language-specific tokenization but also underscores the influence of additional factors, such as model architecture and parameter optimization, in shaping overall performance.

\texttt{llama-3.1} achieved an MMLU score of 70.42\%, supported by a reasonable TR \% (45.80\%), but its lower Pure \% (30.91\%) reveals limitations in capturing the finer-grained morphological details of Turkish. In contrast, \texttt{Qwen2.5}, the smallest model with 7.6B parameters, achieved the lowest MMLU score (61.68\%) and TR \% (40.33\%), highlighting the challenges of maintaining linguistic fidelity within a smaller parameter budget. However, its relatively compact vocabulary size (151,665 tokens) and efficient processing time (3.31 seconds) suggest a trade-off favoring computational efficiency.

\begin{figure}[h!]
    \centering
    \includegraphics[width=1\textwidth]{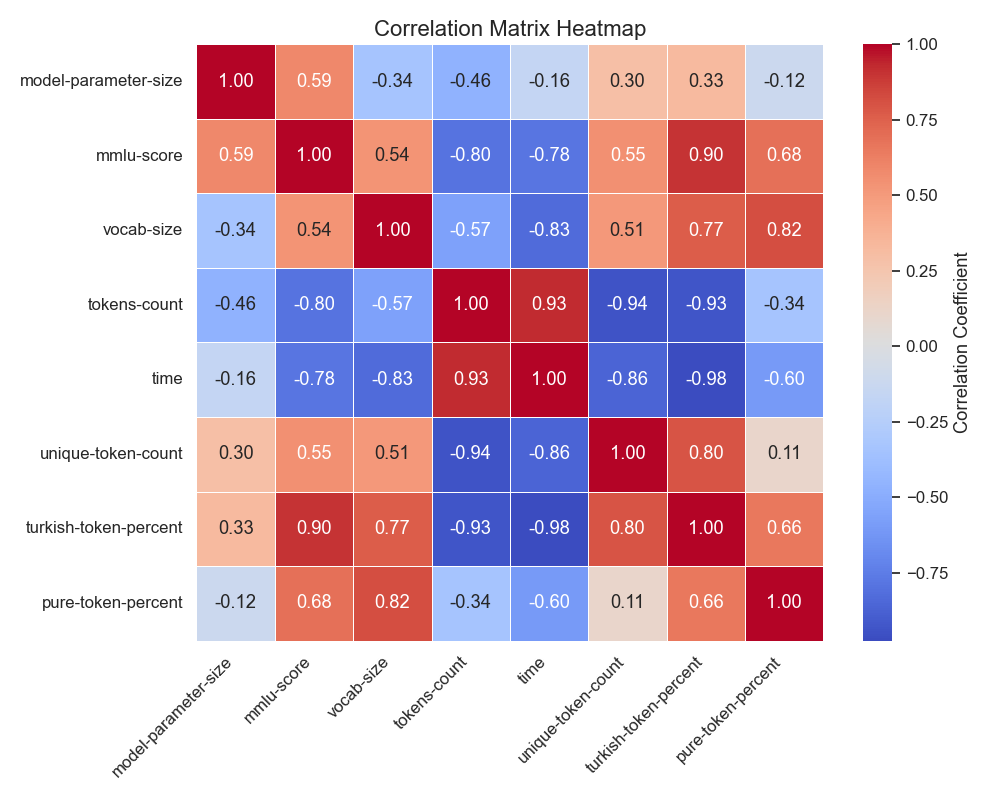}
    \caption{Correlation Matrix Heatmap: Relationships among MMLU Score, Linguistic Metrics (TR \% and Pure \%), Vocabulary Size, and Computational Metrics.}
    \label{fig:correlation_matrix}
\end{figure}

A correlation matrix was computed to analyze the relationships between key metrics, providing deeper insights into how linguistic and computational factors interact. The matrix is visualized in Figure~\ref{fig:correlation_matrix}, which reveals several important patterns. For example, TR \% demonstrated the strongest positive correlation with MMLU scores (\(r = 0.90\)), followed by Pure \% (\(r = 0.68\)), emphasizing the role of linguistic alignment in downstream performance.

\FloatBarrier

The correlation matrix also highlights the nuanced relationship between vocabulary size and linguistic fidelity. Larger vocabularies correlate positively with both TR \% (\(r = 0.77\)) and Pure \% (\(r = 0.82\)), indicating that a sufficiently large vocabulary improves the tokenizer's ability to align with the target language's morphology. However, excessive token counts and processing times exhibit strong negative correlations with these linguistic metrics (\(r = -0.93\) and \(r = -0.60\), respectively), underscoring the inefficiencies introduced by overly granular tokenization.

\begin{figure}[ht]
    \centering
    \includegraphics[width=1\textwidth]{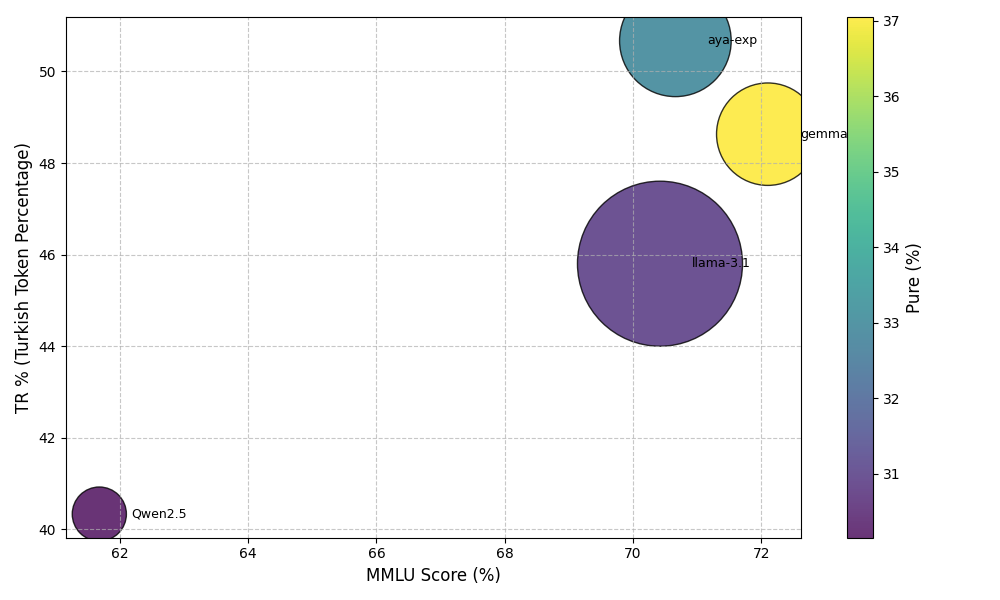}
    \caption{Model Comparison: MMLU vs TR \%, Parameter Size, and Pure \%.}
    \label{fig:model_comparison}
\end{figure}

Figure~\ref{fig:model_comparison} provides a multidimensional comparison of the evaluated tokenizers, plotting MMLU scores against TR \%, with marker size representing parameter count and color encoding Pure \%. Models that achieve high TR \% and Pure \% values tend to better capture the language’s morphological richness, while those optimized for efficiency may sacrifice linguistic fidelity.

\FloatBarrier

These findings affirm the critical role of linguistic alignment, as reflected by TR \% and Pure \%, in shaping downstream performance in morphologically complex languages like Turkish. Models that prioritize language-specific tokenization can outperform larger or more computationally efficient models that lack such adaptation. This highlights the importance of designing tokenization strategies that balance computational efficiency with linguistic fidelity, providing valuable insights for future research and development in NLP.

\section{Conclusion}

This study introduced a comprehensive framework for evaluating tokenization strategies, highlighting the importance of balancing linguistic integrity and computational efficiency in morphologically rich and low-resource languages. By focusing on metrics such as token purity, Turkish Token Percentage (TR \%), processing time, and vocabulary size, the evaluation revealed the significant role of tokenization in shaping downstream model performance. Using the TR-MMLU dataset, which provides a rigorous benchmark for large language models, the analysis demonstrated that linguistic alignment is critical for achieving robust results in morphologically complex languages like Turkish.

The findings revealed that parameter size alone is not a definitive predictor of performance. For example, \texttt{gemma-2} (27.2 billion parameters) outperformed the larger \texttt{llama-3.1} (70.6 billion parameters) in TR-MMLU benchmarks, achieving higher scores for both TR \% and Pure \%, underscoring its superior ability to capture Turkish-specific morphological structures. Conversely, smaller models like \texttt{Qwen2.5}, while computationally efficient, exhibited lower linguistic fidelity due to their limited alignment with Turkish vocabulary and grammar. These results highlight the necessity of language-specific tokenization strategies to optimize performance for morphologically rich languages.

In addition to these findings, the study evaluated several tokenizers under development, such as \texttt{AhmetSemih/tr\_tokenizer} and \texttt{aliarda/turkish\_tokenizer}, which achieved remarkable initial results. These tokenizers demonstrated high TR \% and Pure \%, reflecting their ability to generate tokens aligned with Turkish linguistic structures. However, these initial developments represent only the foundational stage, with substantial potential for further improvements through the integration of advanced morphological analysis, unsupervised learning techniques, and domain-specific adaptations.

The implications of these findings extend beyond Turkish NLP. Linguistically informed tokenization strategies can enhance performance across a wide array of applications, particularly in low-resource settings where linguistic integrity is paramount. For example, in machine translation and sentiment analysis, preserving morphological and syntactic structures can significantly improve accuracy. Similarly, in specialized domains such as healthcare or legal contexts, domain-specific tokenizers can align with specialized terminologies, boosting the precision of text classification and information retrieval tasks.

Future research will focus on iterative refinements of tokenization strategies, including dynamic token generation tailored to downstream tasks and domain-specific requirements. Expanding the framework to evaluate tokenization performance across other morphologically rich languages, such as Finnish or Hungarian, and conducting cross-linguistic comparisons will provide deeper insights into the universal and language-specific aspects of tokenization.

In summary, this study establishes a new standard for evaluating tokenization strategies by combining linguistic fidelity and computational efficiency metrics. It demonstrates that tailored tokenization strategies can enable even smaller or less-optimized models to excel in morphologically complex settings. By advancing these strategies, this research aims to foster the development of robust, linguistically informed tokenizers that enhance the quality and applicability of large language models across multilingual and domain-specific NLP tasks.

\bibliographystyle{unsrt}
\bibliography{tokenizer}

\end{document}